\documentclass[11pt]{article}

\usepackage[preprint]{acl}
\usepackage{tikz}
\usetikzlibrary{positioning}
\usepackage{graphicx}
\usepackage{booktabs}
\usepackage{amsmath}
\usepackage{caption}

\usepackage{booktabs}
\usepackage{tabularx}
\usepackage{float}
\usepackage{multirow}
\usepackage{booktabs}

\usepackage{adjustbox}

\usepackage{times}
\usepackage{latexsym}

\usepackage[T1]{fontenc}
\usepackage[utf8]{inputenc}

\usepackage{microtype}

\usepackage{inconsolata}

\usepackage{graphicx}

\title{RCMN: Understanding Misleadingness in Influential Public Discourse}

\author{Peiling Yi \\
  School of Computer Science and Mathematics \\
  Faculty of Engineering, Computing and the Environment \\
  Kingston University London, United Kingdom \\
  \texttt{p.yi@kingston.ac.uk}
}
\begin{document}

\maketitle
\begin{abstract}

Influential public discourse shapes public beliefs and can also mislead, not only through what is stated, but also through how information is framed, omitted, contextualised, and communicated. Yet less research has focused on how such misleadingness arises and shapes the interpretations formed by readers. To address this gap, we introduce Reader-Centric Misleadingness Understanding (RCMN), a framework that operationalises misleadingness through five dimensions: misleading mechanism, likely reader interpretation, evidence-warranted interpretation, emotional arousal, and communicative intent. Based on this framework, we construct an evidence-grounded dataset of influential public discourse. Empirical findings show that misleadingness is diverse and extends well beyond fabrication, with unsupported inference, exaggeration, and omission among the prevalent mechanisms, and is frequently associated with heightened emotional arousal and distortive communicative intent. Moreover, we investigate whether lightweight claim-and-context representations retain sufficient cues for understanding reader-centric misleadingness without access to richer contextual, evidential, and multimodal information. Evaluation across five recent generative foundation models shows that reader-level interpretations can often be recovered from such limited representations, whereas identifying how misleadingness is produced remains considerably more challenging. These findings highlight the potential of lightweight representations for scalable misleadingness analysis, while reliable understanding of misleading mechanisms continues to require richer contextual and evidential grounding.

\end{abstract}

%Rather than simply increasing the amount of retrieved context, our method aims to identify and prioritise high-quality contextual information while suppressing irrelevant or misleading content. Experimental results are expected to provide new insights into how context influences LLM reasoning in misinformation detection and how adaptive retrieval strategies can improve robustness and reliability. 

%\textcolor{blue}{[Findings:]}We conduct controlled experiments across multiple context levels and context types, including relevant, ambiguous, and misleading information, using both vanilla LLM prompting and retrieval-augmented setups. Our results reveal a consistent non-linear pattern: moderate amounts of high-quality context yield the best performance, while excessive or noisy context leads to increased misclassification, particularly under misleading conditions.
%\endab

\section{Introduction}

Influential public discourse refers to publicly circulated communication that has substantial visibility, prominence, or relevance to public debate and can shape how audiences understand issues of societal concern\cite{druckman2001limits}. The statement plays a central role in shaping how people understand political, social, economic, and other public-interest issues\cite{mccombs1972agenda,entman1993framing,scheufele2007framing}. Such discourse does more than communicate isolated facts: it can frame which aspects of an issue receive attention, establish causal explanations, attribute responsibility, emphasise particular risks or consequences, and influence how audiences interpret subsequent information\cite{chong2010dynamic}. As these messages are increasingly amplified and recirculated through news platforms and social media, their influence may extend well beyond the original communication, contributing to broader public narratives and potentially affecting attitudes, trust, and decision-making\cite{vosoughi2018spread,lazer2018science}.

\begin{figure}[t]
    \centering
    \includegraphics[width=\linewidth]{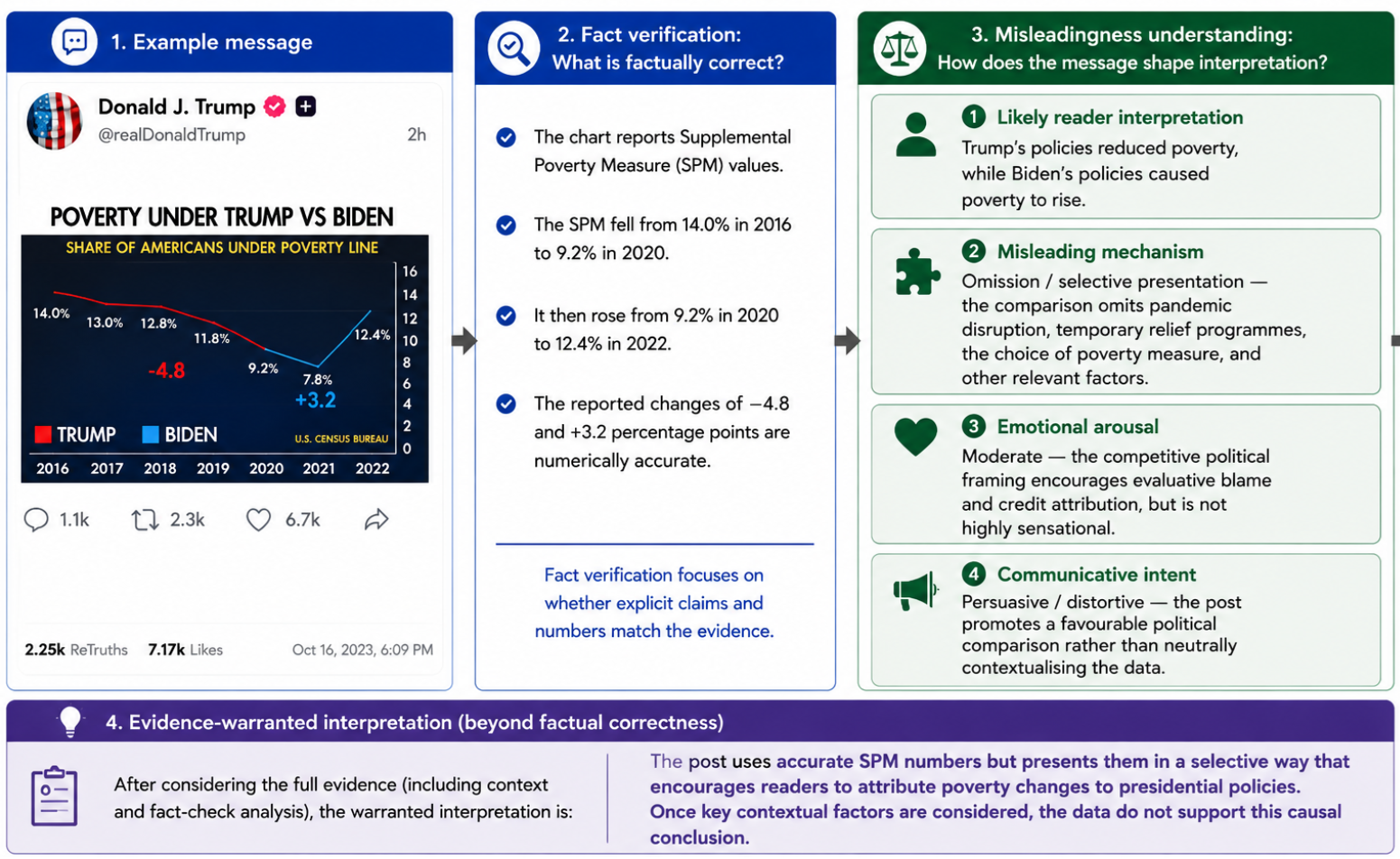}    \caption{From fact verification to misleadingness understanding. All claims, numerical values, and contextual explanations presented in the figure are adapted from the FactCheck.org analysis by \cite{gore2023trump}.}
    \label{fig:misleading-example}
\end{figure}

Given the societal importance of such discourse, its potential to mislead is particularly consequential \cite{ecker2022psychological}, posing a persistent challenge to public knowledge and representative democracy. Crucially, grounded information can lead to a distorted understanding when it is selectively presented, framed, or stripped of relevant context. As illustrated in Figure \ref{fig:misleading-example},  
\textit{At the factual-verification level}, the post reports numerically accurate changes in the Supplemental Poverty Measure (SPM). \textit{At the misleadingness-understanding level}, the same message encourages readers to interpret these changes as evidence that Trump’s policies reduced poverty whereas Biden’s policies caused it to rise. This interpretation is promoted through omission and selective presentation. The comparison gives insufficient attention to the COVID-19 pandemic, temporary economic-relief programmes, the choice of poverty measure, and other relevant socioeconomic factors . Its competitive political framing also produces moderate emotional arousal by encouraging blame-and-credit attribution and serves a persuasive communicative intent rather than a purely informative one. %\textit{Therefore, understanding misleadingness requires moving beyond factual verification to model how information is presented, what interpretations it encourages, and what communicative effects it may produce}.

However, computationally understanding misleadingness remains particularly challenging in
Influential Online Public Discourse. \textit{1) Beyond Veracity-Centred Modelling.} Existing approaches predominantly focus on verifying the factual accuracy of individual claims, while online discourse often derives its persuasive or misleading effect from how claims are framed, combined, contextualised, and presented to audiences\cite{budak2024misunderstanding,pasquetto2024misinformed}. \textit{2) Difficult Operationalisation of Reader-Centric Misleadingness.} Capturing misleadingness in online discourse requires characterising mechanisms of distortion, likely reader interpretations, emotional arousal, and communicative intent. These dimensions depend on substantial contextual judgement and domain knowledge, making reader-centric characterisation considerably more resource-intensive and complex than conventional claim-level assessment.\cite{gabriel2022misinfo,ni2024afacta,modzelewski2026malicious}. \textit{3) Expensive Contextual and Multimodal Reasoning.}
Assessing misleadingness may require recovering rapidly evolving background context and reasoning across text, images, video, or audio to identify omission, emphasis, exaggeration, or recontextualisation. Retrieving and processing such distributed evidence increases computational and latency costs, limiting scalable and real-time analysis \cite{chakraborty2023factify3m,xie2025fire}.

To address these challenges, we propose Reader-Centric Misleadingness Understanding (RCMN), which conceptualises misleadingness in terms of the understanding a message is likely to induce in readers and comprises three complementary components. \textit{1) RCMN Taxonomy.} To move beyond veracity-centred modelling, we characterise misleadingness along five complementary dimensions: misleading mechanisms, likely reader interpretations, Evidence-Warranted Interpretation, emotional arousal, and communicative intent. \textit{2) RCMN Dataset for Online Influential Discourse.} To operationalise these reader-centric dimensions in real-world discourse, we construct RCMN from publicly circulated messages that attracted professional fact-checking attention and are often associated with public figures, organisations, and significant political or social events. By integrating original source materials, contextual information, and expert fact-checking evidence, the dataset provides structured representations of reader-centric misleadingness for fine-grained analysis and model supervision. \textit{3) RCMN Benchmark.} To reduce the cost of contextual and multimodal reasoning at inference time, we benchmark three recent open-source models (Qwen3-VL-8B, DeepSeek-V4-Flash, and Gemma-4-12B) alongside two closed-source models (GPT-5.6 Sol and Claude Fable 5), examining whether they can recover reader-centric misleadingness cues from substantially lower-cost claim-and-context representations without requiring full evidential retrieval or complete multimodal processing.

The empirical findings reveal four main patterns. 
\textit{First}, misleadingness in influential public discourse extends well beyond outright fabrication: most cases arise through unsupported inference, exaggeration, omission, miscontextualisation, or misattribution, demonstrating that factual verification alone is insufficient to capture how messages may mislead readers. 
\textit{Second}, misleading cases are more frequently associated with high emotional arousal and distortive communicative intent, although arousal alone is not sufficient to indicate misleadingness. 
\textit{Third}, in non-misleading cases, the interpretation encouraged by the message generally aligns closely with the evidence-warranted interpretation, supporting interpretive divergence as a useful basis for operationalising misleadingness. 
\textit{Finally}, benchmarking five recent large language and multimodal models reveals a clear asymmetry in recoverability: likely reader interpretations and broad affective and communicative cues can often be recovered from lightweight claim-and-context representations, whereas identifying the precise mechanism of misleadingness remains substantially more difficult and dependent on richer contextual and evidential information, particularly when misleadingness arises from omitted or displaced information.

\textit{Our contribution:} RCMN reframes misleadingness as a structured reader-centric language-understanding problem that captures how misleadingness is produced, what interpretation a message encourages relative to the available evidence, and which affective and communicative signals shape that interpretation. By unifying these dimensions within an evidence-grounded taxonomy, dataset, and benchmark, RCMN provides a systematic framework for studying misleading communication beyond claim-level veracity and for evaluating how effectively five recent generative foundation models can recover reader-centric misleadingness cues from limited contextual information.

%In this study, we find that, despite covering the period from 2019 to 2025, only 2216 instances could be constructed with full evidence-grounded misleadingness annotations derived from professional fact-checking sources, highlighting the scarcity and high cost of such resources. This further motivates the need for scalable misleadingness understanding. Our experiments provide encouraging evidence that lower-cost claim-and-context representations preserve useful reader-centric signals, although they remain better suited to pre-verification screening and prioritisation than to replacing evidence-grounded verification. These findings suggest a promising direction towards more efficient identification of potentially misleading public communication before it spreads widely.

\section{Related work}\label{sec:related work}

In this section, we review related work from two perspectives: benchmark datasets and methodological approaches, tracing the progression from factual verification towards reader-centric understanding of misleadingness. Table~\ref{tab:dataset_comparison}  summarises representative datasets and their expanding task scope.

\begin{table*}[t]
\centering
\caption{Progression of representative datasets from fact verification to
reader-centric misleadingness understanding.}
\label{tab:dataset_comparison}
\scriptsize
\setlength{\tabcolsep}{3.5pt}
\renewcommand{\arraystretch}{1.15}

\begin{tabularx}{\textwidth}{
    p{1.75cm}
    p{0.62cm}
    p{2.25cm}
    p{2.75cm}
    X}
\toprule
\textbf{Dataset} &
\textbf{Year} &
\textbf{Modality} &
\textbf{Primary task} &
\textbf{Labels or outputs} \\
\midrule

% =====================================================

\multicolumn{5}{l}{
\textbf{I. Fact verification and fake-news detection}
} \\

LIAR \cite{wang2017liar}
& 2017
& Text
& Claim-veracity classification
& Six-level truthfulness labels: \textit{pants-fire}, false, barely true,
half true, mostly true, and true. \\

FEVER \cite{thorne2018fever}
& 2018
& Claims and textual evidence
& Evidence-based fact verification
& Supported, refuted, and not enough information. \\

MultiFC \cite{augenstein2019multifc}
& 2019
& Text, evidence, and metadata
& Multi-domain claim verification
& Heterogeneous veracity ratings collected from multiple fact-checking
organisations. \\

FakeNewsNet \cite{shu2020fakenewsnet}
& 2020
& News, social context, and metadata
& Fake-news detection
& Real and fake labels with user-engagement and propagation information. \\

Fakeddit \cite{nakamura2020fakeddit}
& 2020
& Text, image, metadata, and comments
& Multimodal fake-news classification
& Binary, three-way, and six-way fine-grained labels. \\

Factify / Factify 2
\cite{mishra2022factify,suryavardan2023factify}
& 2022--23
& Claims, images, and evidence
& Multimodal fact verification
& Support, refute, insufficient-evidence, and related verification
categories. \\

MuMiN \cite{nielsen2022mumin}
& 2022
& Claims, images, articles, and social graphs
& Multilingual misinformation detection
& Claim-veracity labels linked to multilingual social-network data. \\

MOCHEG \cite{yao2023end}
& 2023
& Claims, images, and web evidence
& Multimodal fact-checking and explanation
& Veracity labels, multimodal evidence, and natural-language explanations. \\

VeriTaS \cite{rothermel2026veritas}
& 2026
& Textual and audiovisual claims
& Dynamic multimodal fact-checking
& Standardised verdict dimensions, retrieved original media, and textual
justifications. \\
\midrule
% =====================================================

\multicolumn{5}{l}{
\textbf{II. Out-of-context and multimodal misalignment detection}
} \\

NewsCLIPpings \cite{luo2021newsclippings}
& 2021
& Image and caption
& Out-of-context misinformation detection
& Pristine and semantically mismatched image--caption pairs. \\

VERITE \cite{papadopoulos2024verite}
& 2024
& Image, caption, and external context
& Real-world out-of-context detection
& Truthful, miscaptioned, and out-of-context image--caption pairs. \\

5Pils-OOC \cite{tonglet-etal-2025-cove}
& 2025
& Image, caption, and retrieved evidence
& Out-of-context detection and image contextualisation
& Accurate/out-of-context captions; predicted original image context and caption veracity. \\

% =====================================================
\midrule
\multicolumn{5}{l}{
\textbf{III. Towards misleadingness understanding}
} \\

M4FC \cite{geng2025m4fc}
& 2025
& Images, multilingual claims, and metadata
& Multitask multimodal fact-checking
& Fake image: authentic / manipulated-or-fake; location verification: whether candidate location is consistent; verdict: true / false. \\

MM-Misleading \cite{li2026s}
& 2026
& News image, headline, and article
& Misleading-omission detection and correction
& misleading / non-misleading based specifically on misleading omission \\

\midrule
\multicolumn{5}{l}{\textbf{IV. Reader-centric misleadingness understanding}}
\\
\textbf{RCMN (Ours)} & 2026 &
Claims, contextual metadata, multimodal sources/links, fact-check metadata/links, and structured evidence &
Reader-centric misleadingness understanding &
Mechanism: fabrication/alteration, miscontextualisation, omission/selective presentation, misattribution, exaggeration/quantitative distortion, unsupported inference; Arousal: low, moderate, high; Intent: informative, persuasive, distortive; Likely reader interpretation; Evidence-warranted interpretation.\\
\bottomrule
\end{tabularx}

\end{table*}

\subsection{Datasets and Task Evolution}

Early misinformation benchmarks largely operationalised the problem through veracity or factuality prediction. LIAR \cite{wang2017liar} introduced fine-grained veracity labels, FEVER \cite{thorne2018fever} classified claims as supported, refuted, or lacking sufficient evidence, and MultiFC \cite{augenstein2019multifc} extended evidence-based verification to naturally occurring claims collected across multiple fact-checking organisations. FakeNewsNet \cite{shu2020fakenewsnet} broadened this setting by incorporating news content, social context, and spatiotemporal information.  Later resources extended misinformation detection and fact-checking to multimodal settings. Fakeddit \cite{nakamura2020fakeddit} provides large-scale text--image examples for multimodal fake-news detection, while Factify \cite{mishra2022factify}, Factify~2\cite{suryavardan2023factify}, MuMiN \cite{nielsen2022mumin}, MOCHEG \cite{yao2023end}  and VeriTaS \cite{rothermel2026veritas} broaden the task space through multimodal fact verification, multilingual and social-context modelling, evidence use, and explanation generation.

A further line of work addresses out-of-context misinformation, where authentic media become misleading through reuse or mismatched contextualisation. NewsCLIPpings \cite{luo2021newsclippings} and VERITE \cite{papadopoulos2024verite} benchmark the detection of out-of-context or miscaptioned image--text pairs, while COVE \cite{tonglet-etal-2025-cove} explicitly reconstructs aspects of an image's original context before assessing caption veracity. These studies importantly demonstrate that misleadingness can arise without media fabrication; These studies importantly demonstrate that misleadingness can arise without media fabrication;

Recent work has begun to move beyond conventional fact verification towards modelling how context, communicative intent, and selective presentation can shape misleading interpretations. M4FC \cite{geng2025m4fc} further incorporates claimant-intent prediction, image contextualisation, location verification, and verdict prediction. MM-Misleading \cite{li2026s} compares preview-supported and article-supported interpretations to identify omission-based cases in which factually compatible news previews nevertheless induce misleading interpretations.

However, these approaches still capture only part of the broader phenomenon of reader-level misleadingness. 

\subsection{Methods}

Reader-centric misleadingness detection shifts the modelling objective beyond factual correctness towards understanding how content, context, communicative intent, emotion, and reader interpretation interact \cite{yi2026fact}. Existing methods provide several foundations for this direction. Human-centred misinformation research models cognitive, emotional, and behavioural responses to misleading content, including inferred writer intent and potential reader actions \cite{gabriel2022misinfo}, while emotion-aware approaches represent reader perception, emotion categories and intensity, and semantic emotion roles \cite{oberlander2020goodnewseveryone}. More recently, interpretation-aware methods compare the understanding induced by limited content with that supported by fuller evidence to detect omission-based interpretation drift \cite{li2026s}. Explainable fact-checking methods further combine multimodal evidence retrieval, veracity prediction, and natural-language explanation generation \cite{yao2023end}. 

However, these methodological strands remain fragmented \cite{yi2026fact}.

\section {RCMN taxonomy}

In the study, \textit{Misleadingness} refers to the extent to which the interpretation encouraged by a message diverges from the interpretation warranted by relevant evidence and context, regardless of whether its individual statements are literally true or false \cite{rogers2017artful,reboul2021truthfully}. %\textit{Fact-checking}, by contrast, primarily evaluates the factual status of a claim by examining whether it is supported or contradicted by available evidence\cite{reboul2021truthfully,thorne2018fever}. 
Drawing on insights from psychology, linguistics, media theory, and communication studies, \cite{yi2026fact} identify three core dimensions that shape misleadingness: emotional arousal, communicative intent, and context. Information may therefore mislead not only through factual inaccuracy, but also by evoking strong emotions, activating moral responses, signalling persuasive or distortive intent, or presenting otherwise accurate information without sufficient context. Building on existing studies, we characterise RCMN along five complementary dimensions.

\subsection{Misleading Mechanism}
This dimension captures the mechanisms or manipulation techniques through which misleading content is produced \cite{van2026prebunking}.

\begin{itemize}

\item \textbf{Fabrication or alteration:} Content is invented, synthetically generated, or technically modified in a way that changes its meaning or evidential value.

\item \textbf{Miscontextualisation:} Authentic content is presented in an incorrect temporal, spatial, event, or discourse context.

\item \textbf{Omission or selective presentation:} Relevant contextual, qualifying, or contradictory information is omitted, or evidence is selectively presented in a way that favours a particular interpretation.

\item \textbf{Misattribution:} Content, a quotation, statement, or action is incorrectly attributed to a person, organisation, publication, or account.

\item \textbf{Exaggeration or quantitative distortion:} The scale, frequency, certainty, severity, or numerical magnitude of a phenomenon is overstated or otherwise distorted.

\item \textbf{Unsupported inference:} The message encourages a causal, evaluative, predictive, or generalised conclusion that is not sufficiently supported by the available evidence and context, even when the underlying statements may be factually accurate.

\item \textbf{Not misleading:} No misleading mechanism is identified when the message is evaluated against the relevant evidence and context.

\end{itemize}

\subsection{Likely Reader Interpretation \& Evidence-Warranted Interpretation}

These two dimensions allow us to measure \emph{interpretive divergence}: the gap between what a message encourages readers to infer and what the available evidence supports. This divergence provides a basis for assessing the degree of misleadingness. \textit{Likely Reader Interpretation} captures the broader inference or conclusion that a message encourages readers to draw, including interpretations shaped by framing, selective presentation, implication, or omitted context \cite{gabriel2022misinfo}. \textit{Evidence-Warranted Interpretation} captures the interpretation justified by the fact-check analysis and relevant contextual evidence \cite{atanasova2020generating}.

\subsection{Emotional Arousal}
This dimension captures the intensity of the emotional response that the content is likely to evoke \cite{luhring2024emotions}.

\begin{itemize}
    \item \textbf{Low}: calm, neutral, or minimally emotional presentation.
    \item \textbf{Moderate}: noticeable but restrained emotional emphasis.
    \item \textbf{High}: strongly alerting, urgent, alarming, or emotionally
    charged presentation.
\end{itemize}

\subsection{Communicative Intent}
This dimension captures the communicative goal that the message appears to serve \cite{da2021edited}.

\begin{itemize}

\item \textbf{Informative communication:} Primarily aims to provide,
report, describe, or explain information to the reader.

\item \textbf{Persuasive communication:} Primarily aims to influence
readers' beliefs, evaluations, or attitudes toward a person, event,
issue, or position.

\item \textbf{Distortive communication:} primarily steer readers toward an interpretation that is not adequately warranted by the available evidence, through selective, exaggerated, decontextualised,
fabricated, or otherwise misleading presentation.

\end{itemize}

\section {RCMN Datasets}

\begin{figure*}[t]
    \centering
    \includegraphics[width=\linewidth]{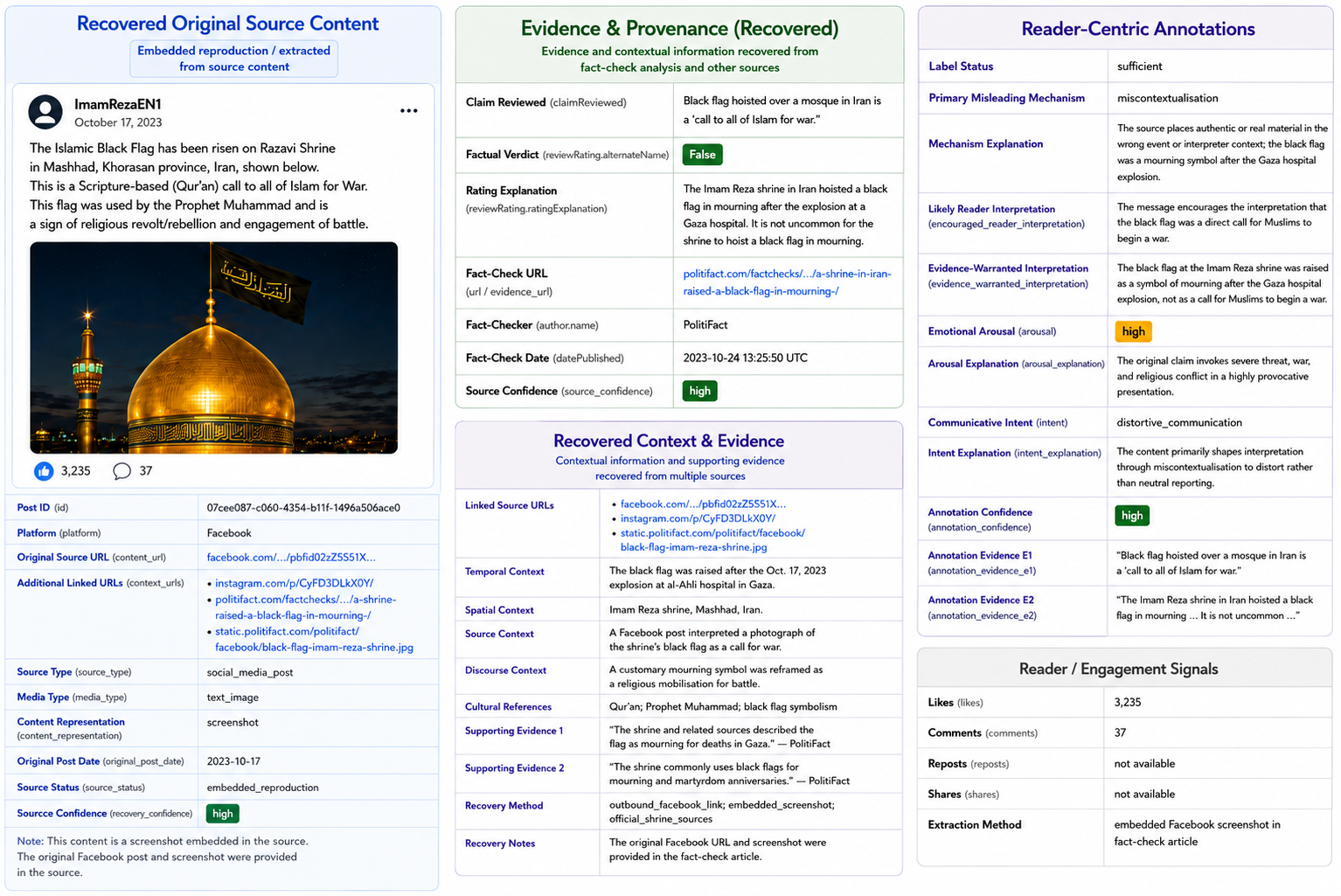} 
    \caption{An example instance from the RCMN dataset}
    \label{fig:dataset-example}
\end{figure*}

Figure~\ref{fig:dataset-example} illustrates the structure of the reader-centric dataset through a representative instance. The dataset is organised into five complementary groups, each serving a distinct purpose. \textit{1) Recovered original source content} records the multimodal content presented to readers, preserving the observable textual and visual information from which an interpretation may be formed. \textit{2) Evidence \& provenance} records the source, provenance, and reference evidence needed to establish what is supported and to assess whether the presented content is misleading. \textit{3) Recovered context and evidence} captures relevant contextual information that is absent or not explicit in the presented content but may materially affect its interpretation. \textit{4) Reader-centric annotations} provide evidence-grounded labels across the five RCMN dimensions, capturing how misleadingness is produced, interpreted, and communicated. \textit{5) Reader engagement signals} provide complementary evidence of how audiences actually respond to and interact with the content.

\subsection{Data Source}

The RCMN dataset is seeded from \textit{Fact Check Insights} \cite{factcheckinsights2026guide}, which aggregates claims drawn from real-world information environments and investigated by independent fact-checking organisations. Rather than treating these records simply as fact-checking examples, we use them as an entry point for identifying and reconstructing potentially misleading public communication. This source is particularly suitable for our study for three reasons: \textit{1)} Fact-checkers tend to investigate claims that have already circulated publicly and attracted social or public attention, providing a useful proxy for potentially influential online discourse in which misleading communication may shape reader understanding. \textit{2)} Its structured metadata and provenance information facilitate tracing claims back to their original communications and recovering the contextual and evidential information needed to analyse how misleadingness is produced and what interpretations it may encourage. \textit{3)} Its coverage across multiple fact-checking organisations exposes RCMN to diverse topics, sources, communicative settings, and forms of misleadingness, supporting the construction of a broad reader-centric taxonomy beyond binary factual-veracity judgements.

%Although many original posts are no longer directly accessible due to deletion, account removal, platform restrictions, or link decay, the associated fact-check articles often preserve substantial information about the original communication through quotations, screenshots, embedded media, source descriptions, and contextual explanations. These materials provide an evidential basis for reconstructing the original message and support evidence-grounded, more reproducible annotation of RCMM. The fact check articles are also embeding  temporal, spatial, source, discourse, and evidential context that is rarely available in conventional misinformation datasets. This combination of structured claim metadata, professional verification, and rich contextual documentation makes Fact-Check Insights particularly suitable for constructing a
%RCMM dataset, even when the original source
%cannot be fully recovered.
%Importantly, Fact Check Insights is used only for
%\emph{candidate identification and evidence recovery}; its fact-check
%ratings are not treated as the target labels of RCMN. Instead, RCMN
%re-annotates the selected cases according to the proposed reader-centric
%taxonomy,

However, Fact-Check Insights exhibit substantial heterogeneity, multilinguality, and sparsity across the full schema. The records span the period from 1970 to 2025 and comprise 260,863 entries across 57 fields, There is no record contain claim source and complete information for all fields. The dataset also demonstrates considerable linguistic and organisational diversity: English accounts for only 7\% of the records, while Filipino is the most represented language. Furthermore, the data include contributions from 1,007 fact-checking organisations and contain 24,869 distinct claim-rating labels, reflecting significant variation in annotation conventions across organisations and languages.

Therefore, we consider Fact-Check Insights only as candidate identification and evidence recovery; its fact-check ratings are not treated as the target labels of RCMN. 

\subsection{Data construct}

\begin{table}[t]
\centering
\caption{Selected core claim fields from Fact-Check Insights.
$^{*}$ denotes fields under \texttt{itemReviewed}, and
$^{\#}$ denotes fields under \texttt{reviewRating}.}
\label{tab:claim_fields}
\small
\begin{tabular}{p{0.38\linewidth} p{0.54\linewidth}}
\hline
\textbf{Field name} & \textbf{Purpose} \\
\hline
\texttt{id}
& Unique identifier for each claim review record. \\

\texttt{claimReviewed}
& Original textual claim being fact-checked. \\

\texttt{datePublished}
& Publication date of the fact-checking article. \\

\texttt{url}
& URL of the fact-checking page. \\

\texttt{author.name}
& Fact-checking organisation or author. \\

\texttt{*.author.name}
& Original claim maker. \\
\texttt{*.author.@type}
& Type of claim maker, such as person or organisation. \\

\texttt{*.name}
& Title of the reviewed item. \\

\texttt{*.datePublished}
& Publication date of the original claim. \\

\texttt{\#.alternateName}
& Veracity label assigned by the fact-checker. \\

\texttt{\#.ratingExplanation}
& Explanation or rationale for the rating. \\
\hline
\end{tabular}
\end{table}

Fact-Check Insights provides only a limited set of structured fields for constructing RCMN, but each record links to a corresponding fact-check article that often preserves or references additional information about the original communication, its context, and the evidence used in the verification process. The fact-check article is therefore used as a recovery gateway to reconstruct the source message and retrieve the contextual and evidential information required for RCMN analysis and annotation.

For each Fact-Check Insights record, eleven base fields are retained, as shown in Table~\ref{tab:claim_fields}, to support record linkage and source recovery. The corresponding fact-check article is then inspected, and its outbound links are followed to recover the original post, article, image, video, advertisement, or archived copy. When the original source is unavailable, the message is reconstructed from screenshots, quotations, embedded media, or descriptions preserved in the fact-check article, with recovery level, content representation, provenance, and confidence recorded accordingly.

\subsection{Annotation}

\begin{figure}[h]
    \centering
    \includegraphics[width=\linewidth]{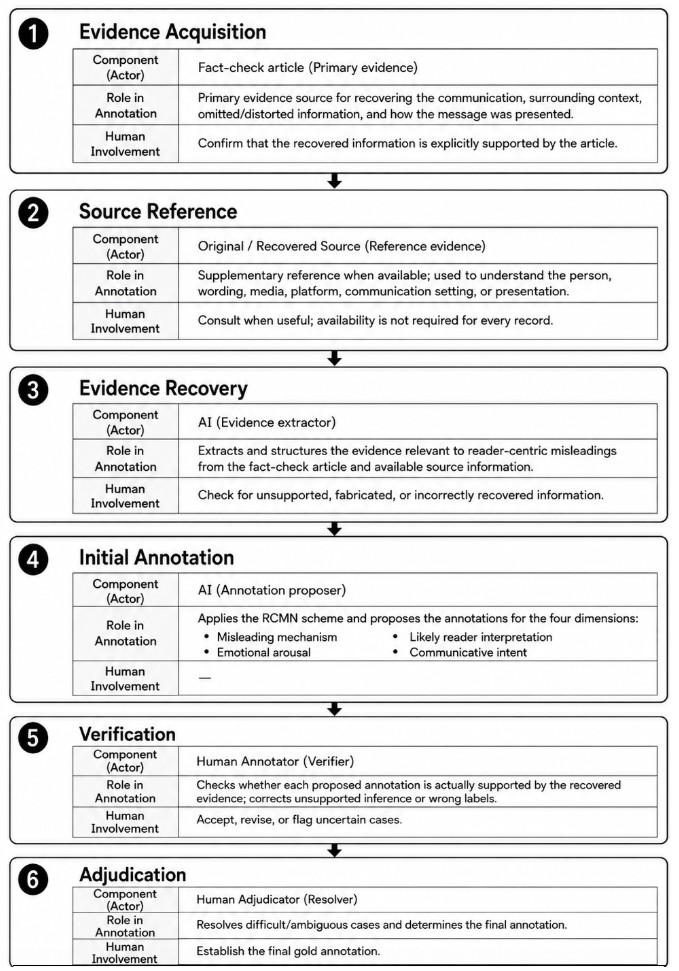} 
    \caption{RCMN dataset annotation pipeline}
    \label{fig:annotaion}
\end{figure}

The annotation pipeline is guided by four principles: evidence grounding, separation of factual veracity from misleadingness, multi-level annotation, and auditability. Accordingly, each annotation and explanation must be supported by explicit evidence and recorded in a form that enables subsequent human verification and adjudication. Importantly, annotations are not derived directly from fact-check verdicts such as \textit{False}, \textit{Mostly False}, or \textit{Half True}; The focus is instead on the communicative process through which a message may encourage a misleading interpretation. The pipeline consists of six stages, as illustrated in Figure~\ref{fig:annotaion}.

\textit{S1) Evidence acquisition.} The complete fact-check article is used as the primary evidence recovery. Relevant information about the reviewed communication, its surrounding context, omitted or distorted information, and how the message was presented is recovered from the article.

\textit{S2) Source reference.} The original or recovered source,such as a social-media post, image, video, quotation, or advertisement, is consulted when available. It serves as supplementary reference evidence for understanding the speaker, wording, media, platform, communication setting, and presentation, but its availability is not required for every record.

\textit{S3) Evidence recovery.} GPT-5.6 Sol, configured with high reasoning effort, was used to extract and structure evidence relevant to reader-centric misleadingness from the fact-check article and available source information. When an article addressed multiple claims, only evidence relevant to the reviewed claim was retained. The recovered evidence was subsequently checked against the source material to remove unsupported, fabricated, or incorrectly attributed information. The recovered evidence was subsequently manually verified against the source material to remove unsupported, fabricated, or incorrectly attributed information.

\textit{S4) Initial annotation.} Apply the RCMN annotation scheme to each instance based on the recovered evidence and context. The annotation considers the relationship between (i) what the communication explicitly states or shows, (ii) the interpretation encouraged by its wording, framing, and presentation, and (iii) the interpretation warranted by the available evidence and context. The model assigns initial labels across the RCMN dimensions and generates an evidence-warranted interpretation as a standardised reference for assessing divergence between the communicated and evidence-supported interpretations.

\textit{S5) Human verification.} Human annotators verify each model-proposed annotation against the recovered source content, evidence, and context. They check whether the misleading mechanism is evidence-supported, whether the likely reader interpretation follows from the presented content, whether the evidence-warranted interpretation reflects the fuller evidence, and whether the arousal and communicative-intent labels are justified by observable cues. The recovered evidence is also checked for incorrect attribution, unsupported additions, or missing material context. Any identified discrepancies are revised, while ambiguous or disputed cases are escalated for further review.

\textit{S6) Adjudication.} Difficult or ambiguous cases are reviewed by a human adjudicator, who resolves disagreements or uncertainty and determines the final gold annotation.

\paragraph{Evidential grounding and annotation quality control.} Sufficient evidence was available for 2,075 of the 2,216 instances (93.6\%), whereas only 101 cases (4.6\%) were judged to have insufficient evidence. Moreover, 2,117 instances (95.5\%) contain direct two-sided evidence, allowing the annotation to be compared with evidence-supported context rather than inferred from the fact-check verdict alone. Annotation reliability was further strengthened through repeated review: 774 instances (34.9\%) received a second check and 323 (14.6\%) received a third check.

\subsection{Empirical Findings}
\begin{table}[!t]
\centering
\caption{Statistics of the deduplicated dataset covering 2019--2025.
Percentages are calculated over all 2,216 instances unless otherwise stated.}

\label{tab:dataset_statistics}

\scriptsize
\setlength{\tabcolsep}{2.5pt}
\renewcommand{\arraystretch}{0.92}

\begin{tabularx}{\columnwidth}{
    >{\raggedright\arraybackslash}X
    r
    r
}
\toprule
\textbf{Category} & \textbf{Count} & \textbf{Share} \\
\midrule

\multicolumn{3}{l}{\textit{Dataset composition}} \\
Unique instances                         & 2,216 & 100.0\% \\
Publication period                       & \multicolumn{2}{r}{2019--2025} \\

double check                  & 774   & 34.9\%$^{\ddagger}$ \\
Three check                & 323   & 14.6\%$^{\ddagger}$ \\
More                 & 1     & 0.05\%$^{\ddagger}$ \\

\addlinespace
\multicolumn{3}{l}{\textit{Annotation status}} \\
Sufficient evidence                      & 2,075 & 93.6\% \\
Article confirms claim / non-misleading  &  63    & 2.84\% \\
Insufficient evidence                   & 101   & 4.6\% \\
\addlinespace
\multicolumn{3}{l}{\textit{Source actor type}} \\
Politician / candidate                    & 973 & 43.9\%$^{\S}$ \\
Political / public organisation           & 92  & 4.2\%$^{\S}$ \\
Media / journalist / commentator          & 57  & 2.6\%$^{\S}$ \\
Social-media / anonymous source           & 688 & 31.0\%$^{\S}$ \\
Other named public / online actor         & 406 & 18.3\%$^{\S}$ \\

\addlinespace
\multicolumn{3}{l}{\textit{Major discourse-domain indicators}} \\
Elections / campaigns                     & 392 & 17.7\%$^{\P}$ \\
Economy / employment                      & 345 & 15.6\%$^{\P}$ \\
Health / public health                    & 318 & 14.4\%$^{\P}$ \\
International affairs / conflict          & 292 & 13.2\%$^{\P}$ \\
Government / public policy                & 217 & 9.8\%$^{\P}$ \\
Crime / public safety                     & 183 & 8.3\%$^{\P}$ \\
Immigration / border                      & 146 & 6.6\%$^{\P}$ \\
Social / cultural issues                  & 141 & 6.4\%$^{\P}$ \\
Climate / environment                     & 73  & 3.3\%$^{\P}$ \\
\addlinespace
\multicolumn{3}{l}{\textit{Fact-checking organisations}} \\
PolitiFact                               & 1,039 & 46.9\% \\
FactCheck.org                            & 540   & 24.4\% \\
FactRakers                               & 276   & 12.5\% \\
Washington Post                          & 226   & 10.2\% \\
Other organisations                      & 135   & 6.1\% \\

\addlinespace
\multicolumn{3}{l}{\textit{Primary misleading mechanism}} \\
Unsupported inference                     & 509 & 24.8\%$^{\dagger}$ \\
Exaggeration / quantitative distortion   & 468 & 22.8\%$^{\dagger}$ \\
Omission / selective presentation        & 361 & 17.6\%$^{\dagger}$ \\
Fabrication / alteration                  & 330 & 16.1\%$^{\dagger}$ \\
Miscontextualisation                      & 230 & 11.2\%$^{\dagger}$ \\
Misattribution                           & 154 & 7.5\%$^{\dagger}$ \\

\addlinespace
\multicolumn{3}{l}{\textit{Emotional arousal}} \\
Low                                       & 305 & 13.8\%\\
Moderate                                 & 634   & 28.6\% \\
High                                    & 1,228 & 55.4\% \\
Not assessable                           & 49    & 2.2\% \\

\addlinespace
\multicolumn{3}{l}{\textit{Communicative intent}} \\
Informative                              & 55    & 2.5\% \\
Persuasive                               & 699   & 31.5\% \\
Distortive                              & 1,406 & 63.4\% \\
Not assessable                          & 56    & 2.5\% \\

\addlinespace
\multicolumn{3}{l}{\textit{Evidence coverage}} \\
Direct two-sided evidence                & 2,117 & 95.5\% \\
No structured evidence                   & 99    & 4.5\% \\
Original-source evidence items           & 2,130 & -- \\
Corrective fact-check evidence items     & 2,404 & -- \\

\bottomrule
\multicolumn{3}{p{\dimexpr\columnwidth-2\tabcolsep\relax}}{
\tiny
$^{\dagger}$ Percentages are calculated over the 2,052 instances with an
assigned mechanism.
$^{\ddagger}$ Percentages are calculated over the 1,098 duplicate groups.
}
\end{tabularx}
\end{table}

Table ~\ref{tab:dataset_statistics} summarises the dataset composition, annotation outcomes, and evidence coverage. More importantly, the dataset reveals several distinctive characteristics of  misleadingness.

\textit{\textbf{F1: Misleadingness represented in the RCMN dataset is strongly embedded in public-facing online discourse.}} Politicians and candidates constitute the largest source group (43.9\%), while elections and campaigns (17.7\%), the economy and employment (15.6\%), public health (14.4\%), and international affairs and conflict (13.2\%) are the most prominent discourse domains.
 
 \textit{\textbf{F2: Misleading mechanisms are diversitiy.}}
The diverse distribution of misleading mechanisms shows that determining whether an individual claim is factually true is insufficient for identifying misleadingness. Most cases do not rely on outright fabrication; instead, they mislead through unsupported inference.

\textit{\textbf{F3: Misleadingness is strongly associated with the way information is communicated}}. More than half of the instances exhibit high emotional arousal, while distortive communicative intent constitutes the largest intent category. Together, these patterns suggest that misleadingness is often encouraged by how information is presented and framed.

\textit{\textbf{F4: Distortive Intent signal misleading}}
All non-misleading cases, 75\% were labelled as persuasive and 23.7\% as informative; 0\% were assigned distortive intent. By contrast, 73.1\% of the misleading cases were labelled as distortive, whereas only 0.3\% were informative. These results suggest that distortive, rather than persuasive, intent may serve as a strong indicator of misleadingness.

\textit{\textbf{F5: Reader-likely interpretation vs evidence-warranted interpretation.}}

A semantic comparison provides further validation. Among cases labelled as non-misleading, the encouraged interpretation shows strong semantic alignment with the evidence-warranted interpretation in nearly all cases, with only one case exhibiting weaker or partial alignment. In contrast, misleading cases exhibit greater divergence between the encouraged and evidence-warranted interpretations. This pattern supports the misleadingness operationalisation: when a message is labelled as non-misleading, the interpretation it encourages is generally consistent with that warranted by the available evidence.

\begin{equation}
\boxed{
\text{Non-misleading}
\;\Rightarrow\;
I_{\text{encouraged}}
\approx
I_{\text{warranted}}
}
\label{eq:nonmisleading-alignment}
\end{equation}

\textit{\textbf{F6: Misleading communication is associated with higher emotional arousal.}}
We observe a clear association between misleadingness and emotional arousal. Among messages with an identified misleading mechanism, 58.3\% exhibit high arousal, compared with only 12.7\% of non-misleading messages. Conversely, 28.6\% of non-misleading messages exhibit low arousal, compared with 13.7\% of misleading messages. The difference in arousal distributions is statistically significant ($\chi^2(2)=51.80$, $p<0.001$), indicating that misleading communication is more frequently associated with emotionally intense presentation. However, the association is small (Cram\'er's $V=0.157$), showing that emotional arousal is associated with, but does not, by itself, determine misleadingness.

\section {Benchmarks}
The RCMN benchmark investigate 
\emph{"To what extent can claim and contextual information preserve sufficient cues for understanding multimodal misleadingness without access to the richer contextual, evidential, and multimodal information used to establish reference annotations?"}

\subsection{Task Definition}

In the study, we formally formulate:

\begin{align}
X_{\text{limited}} = \{&
\text{claim},\,
\text{person},\,
\text{location}, \,
\text{time},\,
\text{event},\nonumber\\
&
\text{source setting},\,
\text{context}
\}, \label{eq:limited}\\[2mm]
X_{\text{full}} = \{&
\text{original source},\,
\text{multimodal content},\nonumber\\
&
\text{fact-check article}, \,
\text{fact-check results},\nonumber\\
&
\text{evidence},\,
\text{source setting},\,
\text{context}
\}
\end{align}

\(X_{\mathrm{full}}\) is used to establish the gold annotation 

\[
Y\{X_{\mathrm{full}}\ \}=
\{Y_m,Y_r,Y_e,Y_i\},
\]

where \(Y_m\) denotes the misleading mechanism, \(Y_r\) the likely reader interpretation, \(Y_e\) the emotional-arousal label, and \(Y_i\) the communicative intent.

At inference time, the evaluated model does not receive the complete evidence used to construct the annotation. Instead, it is provided with a lower-cost representation $X_{\mathrm{limited}}$, a sample shown in Table \ref{tab:rcmn_input}. The model is therefore required to estimate

\[
\hat{Y}=f(X_{\mathrm{limited}})
\]

The objective is for $\hat{Y} \approx Y$.

\begin{table}[t]
\centering
\caption{Example of the input used in the RCMN benchmark.}
\small
\label{tab:rcmn_input}
\begin{tabularx}{\linewidth}{X}
\toprule
\textbf{Input} \\
\midrule
\textbf{Claim/Message:} Credit card debt is above \$1 trillion for the first time ever. 

\textbf{Person/Organisation:} Jim Justice.

\textbf{Location:} United States.

\textbf{Time:} 14 August 2023.

\textbf{Event:} Release of Q2 2023 New York Federal Reserve credit-card balance data.

\textbf{Source setting:} X social-media post.

\textbf{Context:} West Virginia Gov. Jim Justice posted the debt figure while criticising Biden’s economic policies, connecting it to “Bidenomics,” the “radical left,” and families relying on credit cards. The post linked to a CNBC story about the debt milestone. \\
\bottomrule
\end{tabularx}
\end{table}

%\subsection{Experimental Design}
%Rather than evaluating only a single
%\(X_{\mathrm{limited}}\) configuration, we construct three progressively
% pre-verification input settings only
%present in $X_{\mathrm{limited}}$. The three settings therefore 
%form a nested information hierarchy, 
%\[
%X_{\mathrm{limited}}\{\text{claim}\} \subset X_{\mathrm{limited}}\{\text{claim},\,\text{context}\} \subset X_{\mathrm{limited}},\]
%allowing us to quantify the marginal contribution 
%of progressively richer context to reader-centric 
%misleadingness understanding

%\paragraph{X_{\mathrm{limited}}\{\text{claim}\}.}
 %This setting measures how much RCMN information can be inferred directly from the linguistic content of the claim, without any additional contextual information.The model receives only the reviewed claim or message.

%\paragraph{X_{\mathrm{limited}}\{\text{claim},\,\text{context}\}}
%This setting tests whether contextualising
%the claim improves misleadingness understanding beyond the claim alone.
%The model receives the claim together with the recovered
%non-verification contextual description from the fact-check article or original source.

%\paragraph{X_{\mathrm{limited}}}
%The third setting corresponds to the complete \(X_{\mathrm{limited}}\)
%representation defined in Section~\ref{sec:task_definition}:
%This setting provides the rich structured contextual metadata provides
%additional signals beyond textual context alone.

\subsection {Models \& Settings}
To support reproducibility and comparison across model families, we evaluate
three open-weight models: {Qwen3-VL-8B-Instruct}
\cite{bai2025qwen3}, DeepSeek-V4-Flash\cite{xu2026deepseek}, and Gemma-4-12B\cite{team2026gemma}, together with 
GPT-5.6 Sol\cite{openai2026gpt56sol} and Claude-fable-5.
 All models are evaluated under a common zero-shot benchmark protocol using the same 2,216 instances, identical non-verification input fields, task instructions, output schema, and evaluation criteria. Model-specific inference settings are standardised where possible: the open-weight models use 4-bit NF4 quantisation, greedy decoding, and a maximum of 350 generated tokens, while the API models use their respective structured-output interfaces with medium reasoning or thinking settings. Although Qwen3-VL-8B-Instruct and gemma-4-12B are multimodal models, no image input is provided in this benchmark; all models are evaluated using the same limited claim-and-context representation.

\subsection{Evaluation}

\paragraph{Classification.}
Misleading mechanism, emotional arousal, and communicative intent are
formulated as multi-class classification tasks. We use \textit{Macro-F1}
as the primary evaluation metric because it gives equal weight to each class and is therefore less affected by class imbalance. We additionally report class-wise F1 scores to examine performance across individual categories.

\paragraph{Likely Reader Interpretation.}

We evaluate generated likely reader interpretations at both lexical and semantic levels. Lexical similarity is measured using \textit{ROUGE-L}. Because semantically equivalent interpretations may differ substantially in wording, we additionally conduct a meaning-level semantic-equivalence evaluation. Each generated interpretation is compared with the reference annotation and classified as \textit{fully equivalent}, \textit{partially equivalent}, or \textit{non-equivalent}. Full equivalence indicates that the same central reader takeaway is preserved; partial equivalence indicates that the main interpretation is retained but an important qualifier, scope, causal relation, or implication is missing or altered; and non-equivalence indicates a substantially different or contradictory interpretation. Empty or malformed reference interpretations are excluded from semantic evaluation. We further report a semantic adequacy score,
\[
S_{\mathrm{sem}} =
\frac{N_{\mathrm{full}} + 0.5N_{\mathrm{partial}}}
{N_{\mathrm{full}} + N_{\mathrm{partial}} + N_{\mathrm{non}}},
\]
which assigns scores of 1, 0.5, and 0 to fully equivalent, partially equivalent, and non-equivalent outputs, respectively. 

\subsection{Results}

Table~\ref{tab:all_class_f1} presents the class-level classification results, while Table~\ref{tab:reader_interpretation_results} reports the generation performance for likely reader interpretation.

\paragraph{R1: Strong semantic recovery despite limited lexical overlap.}
The generation results reveal a clear distinction between lexical similarity and semantic recovery. Although ROUGE-L scores are relatively modest (0.228 - 0.387), 84 - 97\% of generated interpretations are fully semantically equivalent to the reference, with only <=2\% classified as non-equivalent. Claude Fable 5 provides the clearest example: despite achieving the lowest ROUGE-L score (0.228), it obtains the highest full-equivalence rate (97\%) and a semantic adequacy score of 0.98. This indicates that models can recover the intended reader interpretation from limited claim-and-context information even when their wording differs substantially from the reference. The results also demonstrate that lexical-overlap metrics alone can underestimate the quality of reader-interpretation generation.

\paragraph{R2:Recovery of Emotional and Communicative Cues.}
GPT-5.6 Sol achieves the strongest overall performance for both emotional arousal (Macro-F1 = 0.643) and communicative intent (Macro-F1 = 0.607). Its performance is particularly strong for high arousal (F1 = 0.871), persuasive intent (F1 = 0.966), and distortive intent (F1 = 0.959). These results suggest that broad emotional and communicative cues can often be recovered from limited claim-and-context information, even without conducting full evidential verification or processing the original multimodal content.

\paragraph{R3: Challenging on Misleading-mechanism Classification.}
Misleading-mechanism classification remains challenging across most models, with substantially lower Macro-F1 than emotional arousal for four of the five models. Claude Fable 5 achieves the strongest overall mechanism performance (Macro-F1 = 0.520). The \textit{not misleading} class is particularly difficult for most models, with F1 scores ranging from 0.031 to 0.184 for DeepSeek-v4-flash, Qwen3-VL-8B, GPT-5.6 Sol, and Gemma-4-12B, while Claude Fable 5 performs considerably better (0.393). This pattern may reflect the difficulty of making reliable judgments about misleading mechanisms from limited information, particularly when the available cues appear potentially suspicious.

\paragraph{R4: Intermediate Recoverability of Unsupported Inference.}
Unsupported inference presents an interesting intermediate case. Models can sometimes recognise that a claim makes a stronger causal or interpretive conclusion than is directly supported by the supplied context, yet reliable identification still depends on knowing what conclusions the underlying evidence actually warrants. Taken together, these differences suggest that misleading mechanisms vary in their recoverability from low-cost textual cues. Mechanisms with more explicit lexical, numerical, or presentational signals appear more accessible, whereas mechanisms defined by missing, displaced, or externally verifiable information depend much more strongly on additional evidence.

\begin{table*}[!t] \centering \caption{Per-class and Macro-F1 performance across reader-centric misleadingness dimensions. All models are evaluated on the same 2216 shared instances. Bold indicates the best result in each row.} \label{tab:all_class_f1} \begin{adjustbox}{width=\textwidth} \begin{tabular}{llccccc} \toprule \textbf{Dimension} & \textbf{Class} & \textbf{GPT-5.6 Sol} & \textbf{Qwen3-VL-8B} & \textbf{Deepseek-v4-flash} & \textbf{Gemma-4-12B} & \textbf{Claude Fable 5} \\ \midrule \multirow{8}{*}{\textbf{Mechanism}} & Fabrication / alteration & 0.521 & 0.147 & 0.587 & 0.237 & \textbf{0.634} \\ & Miscontextualisation & 0.350 & 0.123 & \textbf{0.523} & 0.284 & 0.508 \\ & Omission / selective presentation & 0.276 & 0.275 & 0.394 & 0.090 & \textbf{0.463} \\ & Misattribution & 0.199 & 0.149 & 0.173 & 0.080 & \textbf{0.414} \\ & Exaggeration / quantitative distortion & 0.586 & 0.341 & 0.480 & 0.173 & \textbf{0.647} \\ & Unsupported inference & 0.567 & 0.113 & 0.102 & 0.433 & \textbf{0.578} \\ & Not misleading & 0.179 & 0.052 & 0.031 & 0.184 & \textbf{0.393} \\ & \textbf{Macro-F1} & 0.383 & 0.170 & 0.333 & 0.206 & \textbf{0.520} \\ \midrule \multirow{4}{*}{\textbf{Arousal}} & Low & 0.288 & \textbf{0.597} & 0.342 & 0.487 & 0.472 \\ & Moderate & \textbf{0.772} & 0.550 & 0.590 & 0.430 & 0.555 \\ & High & \textbf{0.871} & 0.613 & 0.824 & 0.547 & 0.729 \\ & \textbf{Macro-F1} & \textbf{0.643} & 0.502 & 0.550 & 0.488 & 0.586 \\ \midrule \multirow{4}{*}{\textbf{Intent}} & Informative & \textbf{0.525} & 0.147 & 0.231 & 0.213 & 0.306 \\ & Persuasive & \textbf{0.966} & 0.533 & 0.6139 & 0.433 & 0.501 \\ & Distortive & \textbf{0.959} & 0.603 & 0.749 & 0.421 & 0.824 \\ & \textbf{Macro-F1} & \textbf{0.607} & 0.320 & 0.405 & 0.263 & 0.408 \\ \midrule \multicolumn{2}{l}{\textbf{Mean Macro-F1 across dimensions}} & \textbf{0.544} & 0.331 & 0.429 & 0.319 & 0.505 \\ \bottomrule \end{tabular} \end{adjustbox} \end{table*}

\begin{table*}[t]
\centering
\caption{Evaluation of generated likely reader interpretations. 
ROUGE-L measures lexical overlap, while semantic-equivalence evaluation 
measures meaning-level agreement with the reference interpretation.}
\label{tab:reader_interpretation_results}
\begin{adjustbox}{width=\textwidth}
\begin{tabular}{lcccccc}
\toprule
\textbf{Model} &

\textbf{ROUGE-L $\uparrow$} &
\textbf{Full Equiv. $\uparrow$} &
\textbf{Partial Equiv.} &
\textbf{Non-Equiv. $\downarrow$} &
\textbf{Semantic Adequacy $\uparrow$} \\
\midrule

GPT-5.6 Sol &

\textbf{0.387} &
91\% &
8\% &
1\% &
0.95 \\

Qwen &

0.289 &
84\% &
\textbf{14\%} &
2\% &
0.91 \\

DeepSeek &

 0.256  & 93\% & 5\% & 2\% & 0.95 \\

Gemma &

0.314 &
91\%&
7\% &
1\% &
0.95 \\

Claude Fable 5 & 0.228 & \textbf{97\%} & 3\% &\textbf{ <1\%} & \textbf{0.98} \\

\bottomrule
\end{tabular}
\end{adjustbox}
\end{table*}

\subsection{Discussion}

Returning to the benchmark question, our results show that claim-and-context information preserves substantial but incomplete cues for understanding multimodal misleadingness. Across model families, likely reader interpretations, emotional arousal, and communicative intent can often be recovered without access to the richer evidential and multimodal information used to establish the reference annotations. In contrast, identifying the precise mechanism through which misleadingness arises remains considerably less reliable, particularly when it depends on omitted information, displaced context, or evidence external to the message.

These results reveal an important distinction between \emph{contextual signals of misleadingness} and \emph{context-grounded misleadingness understanding}. Lightweight claim-and-context representations can preserve useful interpretive, affective, and communicative signals, but they are not sufficient for reliably determining whether and how a message misleads. Such judgements often require richer contextual and evidential information about what is absent, how the message relates to its original setting, and what interpretation is warranted by the available evidence. Claim-and-context representations are therefore best viewed as a low-cost basis for preliminary analysis and targeted retrieval, rather than a replacement for full contextual and multimodal reasoning.

\section {Limitations}
This study has several limitations that should be considered when interpreting the findings.

\paragraph{Limited representation of non-misleading communication.}
The dataset is naturally biased towards disputed or potentially misleading claims. As a result, only a small proportion of RCMN instances are labelled as \textit{not misleading}. This imbalance limits the benchmark's ability to evaluate how reliably models distinguish misleading communication from ordinary, evidence-compatible communication, and may partly contribute to the low F1 scores observed for the \textit{not misleading} class. Future work should incorporate a larger and more diverse set of non-misleading controls.

\paragraph{AI-assisted annotation and interpretive subjectivity.}
The annotation process combines AI-assisted evidence recovery and initial annotation with human verification and adjudication. This design improves scalability and auditability, but reader-centric dimensions such as likely interpretation, emotional arousal, and communicative intent remain partly interpretive. Human verification reduces unsupported AI inference, but it cannot eliminate disagreement about how different readers may understand the same communication. The resulting labels should therefore be interpreted as evidence-grounded reference annotations rather than deterministic representations of every possible reader response.

\paragraph{Reconstruction rather than complete multimodal preservation.}
The original post, image, video, or other media is not recoverable for every instance. In such cases, the communication is reconstructed from the fact-check article and associated provenance information, with the original source used as supplementary reference when available. 

\section{Conclusion}

This work introduces RCMN, a reader-centric framework, evidence-grounded dataset, and benchmark for understanding misleadingness in influential public discourse beyond claim-level factuality. RCMN captures how messages become misleading, what interpretations they encourage relative to available evidence, and the affective and communicative signals that shape those interpretations.

Our findings show that reader-centric misleadingness contains both readily observable communicative cues and deeper evidence-dependent mechanisms. This distinction points to a promising direction for future research: developing adaptive models that use lightweight contextual signals for initial assessment while selectively retrieving richer contextual, evidential, or multimodal information when deeper verification is required. Such approaches could support more scalable and reliable understanding of how and why influential public discourse may mislead.

\section{Ethics Statement}

RCMN is constructed from publicly circulated discourse and professionally reviewed source material. The annotation process combines AI-assisted evidence recovery and initial annotation with human verification and adjudication. Because the dataset may contain sensitive or politically salient public communication, the current study focuses on research use and does not infer private attributes or intentions beyond the communicative signals defined in the annotation framework.

\bibliography{RCMM}

\appendix

\end{document}